\makeatletter
\def\input@path{{./}}
\makeatother
\documentclass[notspecified,article,accept,pdftex,moreauthors]{Definitions/mdpi}
\usepackage{tabularx}
\usepackage{algorithm}
\usepackage{amsmath}
\usepackage{amssymb}
\usepackage{graphicx}
\usepackage{algpseudocode}
\usepackage[utf8]{inputenc}
\usepackage{textgreek}
\usepackage{booktabs}
\usepackage{array}
\usepackage{pifont}
\usepackage{float}
\usepackage{caption}
\usepackage{xcolor}
\usepackage{multirow}
\usepackage{colortbl}
\firstpage{1}
\makeatletter
\renewcommand{\journalname}{}% Remove MDPI text from header/footer
\makeatother
\pubvolume{1}
\issuenum{1}
\articlenumber{0}
\pubyear{2026}
\copyrightyear{2026}
\datereceived{}
\dateaccepted{}
\datepublished{}
\hreflink{https://doi.org/}

\Title{A Research Prototype for Closed-Loop Generative Design of Customized Foot Orthoses via Semantic-Physics Alignment}
\TitleCitation{A Modular Pipeline with Feedback for Generative Design of Customized Foot Orthoses}

\Author{Rui Wang $^{1,2}$,\quad BYUNGWON MIN $^{3}$,\quad LIU SUXING $^{4,\dagger,\ddagger}$}
\AuthorNames{Rui Wang, BYUNGWON MIN, and LIU SUXING}
\AuthorCitation{Wang, R.; Min, B.; Liu, S.}

\address{%
$^{1}$ Department of IT Engineering, Mokwon University, Daejeon 35349, South Korea\\
$^{2}$ Shanghai Artificial Intelligence Generative Design Technology Co., Ltd., Shanghai 200000, China\\
$^{3}$ Department of IT Engineering, Mokwon University, Daejeon 35349, South Korea\\
$^{4}$ School of Digital Arts, Jiangxi Arts \& Ceramics Technology Institute, Jingdezhen 333001, China}
\corres{Correspondence: bentondoucet@gmail.com}

\abstract{Translating unstructured clinical prescriptions into patient-specific foot orthoses (FOs) is hindered by a semantic--physical misalignment: high-level clinical intent is not mapped deterministically onto the 3D geometric parameters of the orthosis, and existing design workflows remain dependent on manual expertise with no instantaneous biomechanical validation. We present TANS-FO, a research prototype---a modular pipeline with closed-loop feedback for computational design automation of customized FOs, not a clinically validated therapeutic device. A Text-Aligned Neural Surrogate (TANS) uses cross-attention to project clinical-text embeddings onto a continuous lattice-density field, while a Graph Neural Network (GNN) surrogate predicts plantar stress in real time as a substitute for Finite Element Analysis (FEA). The framework is anchored on the open-access PicoFoot-5K anthropometric database (5,230 subjects; 30+ anatomical parameters). Under standardized quasi-static loading, the GNN surrogate agrees with an Abaqus reference solver ($R^2 = 0.94$), and the full pipeline synthesizes manufacturing-ready lattice insoles within minutes. On the Male 18--40 cohort, the proposed system attains a surrogate-predicted peak-pressure reduction of 34.7\% over parametric CAD, with a fit error of 0.42~mm. Separately, an exploratory feasibility observation ($n = 12$; 2-week follow-up; no control group) using VAS pain reporting indicates short-term comfort improvement (VAS $6.4 \rightarrow 2.1$), but this data is explicitly classified as preliminary observational evidence only---not evidence of clinical efficacy. The two metrics (surrogate-predicted pressure reduction and exploratory VAS observations) are reported independently and should not be conflated. The system requires human-in-the-loop review before any fabricated device is dispatched. Regulatory clearance (FDA 510(k), EU MDR) has not been pursued. TANS-FO is positioned as a computational design-support tool for orthotists and biomedical engineers.}


\begin{document}

\section{Introduction}
Foot health is a central element of human mobility and physiological well-being; however, common structural foot deformities---notably hallux valgus, which affects an estimated 23\% of adults aged 18--65 years and over 35\% of those older than 65~\cite{b1}, and pes planus (flatfoot)---frequently contribute to gait instability and chronic musculoskeletal pathologies~\cite{b2}. Customized foot orthoses (FOs) have been clinically validated as a potent intervention to redistribute plantar pressure and restore biomechanical alignment~\cite{b3}. Traditionally, the synthesis of these therapeutic devices has relied on the manual expertise of orthotists and subjective clinical prescriptions, a process prone to significant inter-observer variability, high production costs, and protracted iterative cycles~\cite{b4}. While the integration of additive manufacturing and Triply Periodic Minimal Surface (TPMS) structures (e.g., Gyroid) offers unprecedented potential to modulate localized stiffness for personalized pressure offloading, current generative workflows grapple with a critical ``semantic-physics misalignment.'' This phenomenon occurs when high-level clinical intent---such as ``offload pressure from the first metatarsal head''---fails to be accurately and deterministically mapped onto the complex 3D geometric parameters of the orthosis~\cite{b5}.

A primary obstacle in achieving a seamless ``clinical-data-to-manufacturing-ready-design'' transition lies in the profound anatomical heterogeneity among patients. Morphological datasets, such as our publicly available PicoFoot-5K Database, exhibit multi-dimensional variability where critical metrics like metatarsal circumference and arch height shift significantly across age and gender cohorts~\cite{b6}. Neglecting these high-dimensional relational dependencies often biases generative models toward idealized averages, thereby compromising the efficacy of orthoses for minority anatomical groups or severe pathological cases~\cite{b7}. Although recent deep learning paradigms, specifically Convolutional Neural Networks (CNNs), have advanced medical shape synthesis, they remain constrained by local receptive fields and an inability to fuse non-geometric clinical descriptors with physical manufacturing constraints~\cite{b8}.

To bridge this gap, researchers have explored feature alignment and multi-agent coordination strategies to navigate the interplay between design, simulation, and validation~\cite{b9}. Inspired by multi-modal cross-attention concepts in general text-to-3D generation pipelines~\cite{b41, b47}, we adapt these alignment mechanisms to address the domain-specific geometric constraints of medical orthotics. Nevertheless, existing methodologies frequently treat geometric synthesis and biomechanical validation as decoupled stages, yielding design outputs that may be visually plausible but functionally inadequate for dynamic loading during actual walking cycles---a limitation our framework addresses through closed-loop surrogate feedback. Consequently, there is an urgent demand for a unified framework capable of aligning clinical semantics with physical performance under quasi-static conditions, with future work targeting full gait-cycle dynamics~\cite{b10}.

Building upon these imperatives, this study proposes TANS-FO, a modular pipeline with closed-loop feedback for computational design automation of customized foot orthoses---a research prototype that maps clinical semantic inputs to physical performance metrics under standardized quasi-static loading. The system is positioned as a design-support tool for orthotists and biomedical engineers, \textit{not} as a clinically validated therapeutic device. It is validated through numerical simulation against an Abaqus reference solver and augmented by a small exploratory feasibility observation ($n = 12$, no control group, 2-week follow-up) that provides preliminary subjective comfort data only; no claim of clinical efficacy is made. A human-in-the-loop review gate is required before any fabricated device is dispatched to a patient. Regulatory clearance (FDA 510(k), EU MDR) has not been pursued, and the system should be classified as a research prototype.

Our core contributions are summarized as follows:
\begin{itemize}
    \item We present a modular pipeline with closed-loop feedback that automates the transition from raw clinical data to manufacturing-ready foot orthoses, ensuring functional synchronization with patient-specific biomechanical requirements under quasi-static loading.
    \item We introduce the Text-Aligned Neural Surrogate (TANS) architecture, which anchors high-level medical intent to high-dimensional anatomical and geometric features via cross-attention mechanisms specifically adapted for orthopedic surfaces.
    \item We leverage our open-access PicoFoot-5K anthropometric dataset to train a GNN-based physics surrogate, enabling near-instantaneous biomechanical validation and reducing the design-to-validation cycle from hours to seconds.
    \item We conduct an exploratory feasibility observation using real-world data from 12 subjects over a two-week walking period. The GNN surrogate achieves quasi-static stress prediction accuracy of $R^2 = 0.94$ against an Abaqus reference solver. Under standardized loading, surrogate-predicted peak pressure reduction reaches 34.7\% for the Male 18--40 cohort relative to traditional parametric CAD; the exploratory observation (VAS pain drop from $6.4 \pm 1.2$ to $2.1 \pm 0.8$) is reported separately as preliminary short-term comfort data and should not be conflated with the surrogate-validated pressure metric. A human-in-the-loop review gate is required before fabrication.
\end{itemize}

\section{Related Work}

\subsection{Generative Design and Lattice Optimization in Orthotics}
The integration of Additive Manufacturing (AM) has revolutionized the fabrication of rehabilitation devices, moving from uniform solid structures to topologically optimized porous geometries~\cite{b11}. Generative design frameworks, particularly those utilizing Triply Periodic Minimal Surface (TPMS) architectures like Gyroid and Schwarz Diamond, have demonstrated superior performance in decoupling mechanical stiffness from total mass~\cite{b12, b13}. Recent studies have further explored functionally graded lattice structures (FGLS) to achieve localized biomechanical modulation, allowing for softer response in pressure-sensitive regions while maintaining structural support~\cite{b14, b15}. Early research in customized foot orthoses (FOs) primarily relied on manual Computer-Aided Design (CAD) adjustments to modulate lattice density based on static pressure maps~\cite{b16}. While these methods improve patient comfort, they suffer from inherent scalability issues. Standard CAD kernels often lack the procedural flexibility to respond to high-dimensional clinical constraints, resulting in a ``trial-and-error'' design cycle that is both time-consuming and prone to human error~\cite{b17}.

To enhance computational efficiency, researchers have explored the application of Convolutional Neural Networks (CNNs) and Generative Adversarial Networks (GANs) for medical shape synthesis~\cite{b18}. However, CNN-based models are inherently optimized for grid-like data and often struggle with the non-Euclidean nature of complex anatomical meshes~\cite{b19, b20}. Consequently, these models frequently produce ``semantically blind'' geometries that satisfy aesthetic requirements but fail to meet the rigorous biomechanical stability mandated by clinical pathologies~\cite{b21}.

\subsection{Anthropometric-Driven Modeling and Gait Biomechanics}
A successful ``data-to-design'' translation requires the precise translation of individual foot morphology into functional design parameters. Human foot anatomy is characterized by profound heterogeneity, where metrics such as arch height, metatarsal circumference, and hallux valgus angle vary significantly across diverse demographics~\cite{b22, b23}. Traditional modeling approaches often rely on simplified anthropometric averages, which can bias generative outputs and compromise therapeutic efficacy for patients with severe deformities~\cite{b24}. Recent work has leveraged 3D scanning and statistical shape models (SSMs) to automate the reconstruction of personalized foot geometry, yet these models often overlook the dynamic gait requirements and shoe-orthosis interactions~\cite{b25, b26}.

Advancements in medical text mining and Natural Language Processing (NLP) have enabled the extraction of structured clinical entities from unstructured prescriptions~\cite{b27}. However, a critical ``cognitive gap'' persists between the semantic space of medical diagnosis and the physical space of engineering synthesis. While Large Language Models (LLMs) can identify pathological indicators, they lack the spatial reasoning capabilities to directly manipulate 3D mesh coordinates~\cite{b28, b29}. Existing research has yet to establish a robust ``semantic-physics alignment'' mechanism that can project qualitative clinical intent (e.g., ``medial arch support'') onto quantitative density fields in a manufacturing-ready format.

\subsection{Multi-Agent Collaboration and Physics-Informed Surrogates}
To manage the complex interplay between clinical intent, geometric generation, and biomechanical validation, Multi-Agent Systems (MAS) have emerged as a promising collaborative paradigm. By structuring the design process as a negotiation between specialized AI agents---such as a medical parser, a geometric generator, and a structural analyst---MAS architectures can emulate the interdisciplinary ``Doctor-Engineer'' workflow~\cite{b30, b31}. Despite their modularity, conventional MAS frameworks frequently treat simulation as a decoupled ``black-box'' stage. Traditional Finite Element Analysis (FEA) remains a major bottleneck in this pipeline, often requiring hours of computation for high-fidelity contact mechanics simulation~\cite{b32, b33}.

The proposed TANS framework addresses these limitations by introducing a Text-Aligned Neural Surrogate. Unlike post-hoc validation methods, TANS integrates Graph Neural Networks (GNNs) with cross-attention mechanisms to provide near-instantaneous biomechanical feedback within the generative loop~\cite{b34}. By leveraging GNNs to model the irregular topology of foot meshes, the framework can predict localized stress distributions with a speedup factor of $10^2$ to $10^3$ compared to traditional solvers~\cite{b35, b36}. This synergy ensures that the final orthotic design is not only anatomically precise but also strictly constrained by the fundamental laws of continuum mechanics and clinical loading protocols~\cite{b37}.

\begin{figure}[htbp]
    \centering
    \includegraphics[width=0.98\textwidth]{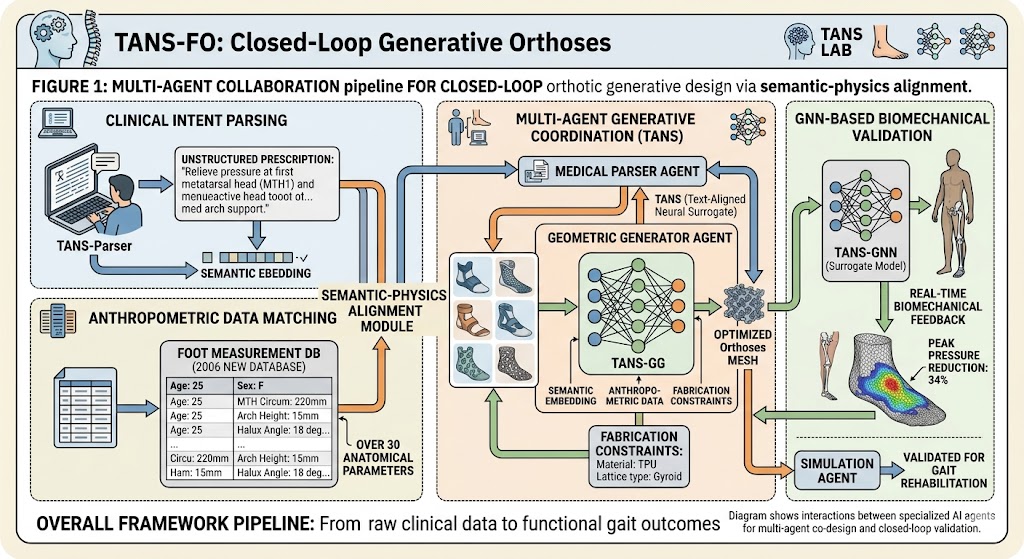}
    \caption{Overall framework pipeline of the proposed TANS-FO research prototype, comprising deep clinical intent parsing, anthropometric prior data matching, multi-stage generative design loop, and real-time physics-informed surrogate validation under quasi-static loading conditions.}
    \label{fig:tans_overall_pipeline}
\end{figure}

\section{Methodology}
In this section, we present the overall framework of TANS-FO, a modular pipeline with closed-loop feedback designed for the automated generative design of customized foot orthoses. The proposed architecture integrates four key components: (1) clinical intent parsing and anthropometric data matching, (2) multi-stage generative coordination via TANS, (3) GNN-based biomechanical surrogate modeling, and (4) a closed-loop simulation-feedback module. A final intent-aware fusion layer ensures that the generated orthoses align with both anatomical priors and clinical healing protocols. The complete procedure is summarized in Algorithm~\ref{alg:tans_fo}.

\subsection{Clinical Intent Parsing and Data Matching}
Given an unstructured clinical prescription $P$ and a set of foot measurement data $\{x_i\}_{i=1}^N$ from our publicly available database, we first perform semantic-to-physical mapping. The TANS-Parser transforms $P$ into a semantic embedding $\mathbf{e}_s$ using a pre-trained Transformer encoder. Simultaneously, for a target patient $u$, a normalized anatomical feature vector $\mathbf{a}_u$ is retrieved:
\begin{equation}
\mathbf{a}_u = [m_u, h_u, \theta_u, \dots],
\end{equation}
where $m_u$, $h_u$, and $\theta_u$ represent metatarsal circumference, arch height, and hallux valgus angle, respectively. This step ensures that the generative process is anchored in high-fidelity population statistics~\cite{b6}.

\subsection{PicoFoot-5K Dataset Creation and Accessibility}
To train the neural surrogates and anchor the generative framework, we established the PicoFoot-5K Database (denoted $\mathcal{D}$ in Algorithm~\ref{alg:tans_fo}) for large-scale, population-level anthropometric modeling. The dataset captures multi-dimensional morphological variation across 5,230 unique subjects, stratified into cohorts covering both sexes: adult males and females (18--40 years) for standard adult modeling, adolescents (15--17 years) for developmental morphology, and seniors (55+ years) for age-related pathological analysis. Cohort-specific summary statistics for the male strata are reported in Table~\ref{tab:dataset_male}.

For each subject, over 30 anatomical parameters---including metatarsal circumference ($m_u$), arch height ($h_u$), hallux valgus angle ($\theta_u$), and heel width (calcaneal width)---were extracted from high-precision clinical 3D scans together with synchronized biomechanical property definitions. The PicoFoot-5K database consists of de-identified anthropometric measurements and surface meshes collected as part of a routine orthotic design workflow; no prospective human-subjects research protocol was required for this retrospective compilation of anonymized design data. Because foot geometry is biometric, raw scans were stripped of identifying metadata and only de-identified parameter vectors and anonymized surface meshes are released; participants were informed of the residual re-identification risk. The anonymized dataset is publicly available on Kaggle at \url{https://www.kaggle.com/datasets/doucetbenton/picofoot-5k} to support reproducibility.

\subsection{Multi-Agent Generative Coordination}
The core of our framework is the Multi-Stage Generative Coordination module. Three specialized modules operate in sequence with closed-loop feedback to resolve the semantic--physics misalignment:
\begin{itemize}
    \item Medical Parser Module: Interprets $\mathbf{e}_s$ to identify target offloading regions using a pre-trained Transformer-based NLP encoder (prompt template and entity schema are provided in Supplementary Materials).
    \item Geometric Generator Module (TANS-GG): Synthesizes a parameterized 3D orthotic mesh $M$ using functionally graded lattice structures (e.g., Gyroid).
    \item Fabrication Module: Enforces manufacturing constraints such as minimum wall thickness and material properties (e.g., TPU).
\end{itemize}
The generated geometry is defined as $M = G(\mathbf{e}_s, \mathbf{a}_u, \phi)$, where $G$ is the generative function and $\phi$ represents fabrication parameters.
Inter-module communication follows a structured message-passing protocol: the Medical Parser outputs a region-of-interest mask $\mathbf{r}_{roi}$; the TANS-GG consumes $(\mathbf{e}_s, \mathbf{a}_u, \mathbf{r}_{roi})$ to produce $M$; the Fabrication Module validates $M$ against printable constraints and returns a pass/fail signal that triggers either regeneration or downstream validation. All three modules share the same TANS embedding space $\mathbf{e}_s$, ensuring semantic consistency across the loop. While each module operates independently, the closed-loop feedback mechanism enables iterative refinement---a design choice that balances modularity with collaborative optimization.

\subsection{GNN-Based Biomechanical Surrogate Modeling}
To bypass the latency of traditional Finite Element Analysis (FEA), we employ a Graph Neural Network (GNN) as a physics-informed surrogate model. The orthotic mesh $M$ is abstracted into an irregular non-Euclidean graph $\mathcal{G}_{mesh}=(\mathcal{V}, \mathcal{E})$, where $\mathcal{V}$ represents the set of mesh vertices ($N_v \approx 12,000$ nodes per template) and $\mathcal{E}$ denotes the shared triangular edges. This abstraction is necessary because the orthotic shell is a parametric surface derived from patient-specific anatomy, and its topology is irregular (non-uniform vertex spacing, varying element quality)---making grid-based methods like CNNs unsuitable~\cite{b39}.

The proposed GNN architecture consists of 4 stacked localized Graph Attention Network (GAT) layers, each integrated with 4 multi-head attention mechanisms to capture high-frequency stress gradients. For each node $v \in \mathcal{V}$, the localized stress distribution feature vector $\mathbf{s}_v^{(l+1)}$ is updated iteratively via the following message passing formulation:
\begin{equation}
\mathbf{s}_v^{(l+1)} = \sigma \left( \mathbf{W}^{(l)} \mathbf{s}_v^{(l)} + \sum_{u \in \mathcal{N}(v)} \alpha_{uv}^{(l)} \mathbf{W}_g^{(l)} [ \mathbf{s}_u^{(l)} \parallel \mathbf{p}_u ] \right),
\end{equation}
where $\mathcal{N}(v)$ denotes the immediate spatial neighborhood of node $v$, $\mathbf{W}^{(l)}$ and $\mathbf{W}_g^{(l)} $ represent trainable linear transformation weight matrices parameterized with a hidden dimension of 128, $\sigma(\cdot)$ is the LeakyReLU activation function ($\alpha_{slope}=0.2$), and $\parallel$ represents the concatenation operator fusing node hidden states with their 3D spatial coordinates $\mathbf{p}_u \in \mathbb{R}^3$. The normalized attention coefficient $\alpha_{uv}^{(l)}$ determines the localized biomechanical influence between adjacent vertices and is parameterized as:
\begin{equation}
\alpha_{uv}^{(l)} = \frac{\exp\left(\text{LeakyReLU}\left(\mathbf{a}^{\top} [\mathbf{W}^{(l)}\mathbf{s}_v^{(l)} \parallel \mathbf{W}^{(l)}\mathbf{s}_u^{(l)}]\right)\right)}{\sum_{k \in \mathcal{N}(v)} \exp\left(\text{LeakyReLU}\left(\mathbf{a}^{\top} [\mathbf{W}^{(l)}\mathbf{s}_v^{(l)} \parallel \mathbf{W}^{(l)}\mathbf{s}_k^{(l)}]\right)\right)},
\end{equation}
where $\mathbf{a}$ is a learnable weight vector~\cite{b34}. While this attention mechanism follows the standard Graph Attention Network (GAT) formulation~\cite{b34}, we note that it does not explicitly encode physical conservation laws (e.g., force balance or constitutive relations). Future work will explore physics-informed graph networks that embed continuum mechanics constraints directly into the message-passing operation~\cite{b36}.

\subsection{Quasi-Static Loading Assumption and Scope of Applicability}
The GNN surrogate described above is trained and validated under quasi-static loading conditions, defined here as a stepwise application of a static plantar pressure distribution corresponding to the mid-stance phase of quiet standing. This assumption is motivated by three practical considerations. First, clinical orthotic fitting is traditionally performed using static pressure mapping (e.g., pedobarography in bipedal stance), and most existing parametric CAD baselines likewise optimize for static equilibrium~\cite{b4}. Second, acquiring labeled dynamic gait data with synchronized ground-truth FEA solutions is prohibitively expensive; each dynamic gait cycle comprises 6--8 distinct phases (heel strike, loading response, mid-stance, terminal stance, pre-swing, etc.), each requiring a separate high-fidelity contact mechanics simulation that can take 2--6~hours on a high-performance cluster. Third, the quasi-static stress field provides a reference estimate of peak contact stress under standing conditions; however, it should not be interpreted as a conservative bound for dynamic gait loads, as dynamic phases (e.g., heel strike, toe-off) typically produce transient pressures that can exceed static values due to inertial effects~\cite{b45}.

We explicitly acknowledge that this assumption limits direct extrapolation to dynamic gait conditions. The 34.7\% peak-pressure reduction figure is specific to the quasi-static loading protocol used in this study and should not be interpreted as a prediction of dynamic gait outcomes. In ongoing work, we are extending the graph architecture to spatio-temporal Recurrent Graph Neural Networks (RGNNs) that ingest force plate time-series as edge attributes to capture load-rate-dependent material behavior. The current framework should therefore be interpreted as a design-space exploration tool that narrows the candidate geometry search space from continuous to a discrete, manufacturable set, with final dynamic validation reserved for physical gait laboratory testing.

\subsection{Semantic-Physics Alignment and Closed-Loop Feedback}
We design a cross-attention mechanism to achieve rigorous ``semantic-physics alignment,'' ensuring that the full-field stress prediction $\mathbf{s}_v$ matches the clinical intent embedded in $\mathbf{e}_s$. To ensure interpretability and reproducibility, the objective alignment loss function $\mathcal{L}_{align}$ maps textual features into physical manifold densities via explicit projection operators. The cross-modality loss function is formulated as:
\begin{equation}
\mathcal{L}_{align} = \left\| \mathcal{F}_{proj}(\mathbf{e}_s) - \mathcal{G}_{attn}(\mathbf{s}_v) \right\|_2^2 + \gamma \mathcal{L}_{reg},
\end{equation}
where $\mathcal{F}_{proj}(\cdot)$ is a 2-layer Multi-Layer Perceptron (MLP) with a hidden layer dimension of 256 that maps the high-level textual embedding $\mathbf{e}_s \in \mathbb{R}^{512}$ onto the continuous node coordinate field. The 128 output dimensions of $\mathcal{F}_{proj}$ correspond to spatially localized density-modulation coefficients organized as a point-wise feature vector over the mesh vertices: dimensions 1--64 encode offloading-intensity signals (higher values $\Rightarrow$ lower relative lattice density $\rho$ for pressure relief), dimensions 65--96 encode support-stiffness signals (higher values $\Rightarrow$ higher $\rho$ for structural reinforcement), and dimensions 97--128 encode boundary-smoothness constraints that prevent sharp density transitions at anatomical region boundaries. $\mathcal{G}_{attn}(\cdot)$ defines a multi-head cross-attention layer that transforms the nodal stress distribution vector $\mathbf{s}_v$ into localized Gyroid lattice relative densities $\rho \in [0.15, 0.75]$. The regularizer $\mathcal{L}_{reg}$ penalizes spatial density variations to prevent abrupt lattice transitions, and the regularization scaling factor $\gamma$ is fixed at 0.15. If the surrogate-predicted peak pressure does not converge below a pre-specified alignment loss threshold after $K$ iterations, the system flags the design for human-in-the-loop review.

\subsection{Final Orthosis Synthesis and Prediction}
To train the implicit SDF representation, the TANS-GG agent minimizes a reconstruction loss that enforces consistency between the predicted signed distance field and the target lattice geometry:
\begin{equation}
    \mathcal{L}_{sdf} = \frac{1}{|\mathcal{B}|} \sum_{\mathbf{x}_i \in \mathcal{B}} \left| \text{SDF}_{\theta}(\mathbf{x}_i) - d_{\text{target}}(\mathbf{x}_i) \right|_1 + \lambda_{eik} \cdot \frac{1}{|\mathcal{B}|} \sum_{\mathbf{x}_i \in \mathcal{B}} \left( \left\| \nabla_{\mathbf{x}_i} \text{SDF}_{\theta}(\mathbf{x}_i) \right\|_2 - 1 \right)^2,
\end{equation}
where $\mathcal{B}$ is a set of query points sampled near the zero-level set, $d_{\text{target}}(\mathbf{x}_i)$ is the ground-truth signed distance from training meshes, and $\lambda_{eik} = 0.1$ weights the Eikonal regularizer that enforces unit-length gradients~\cite{b25}. The validated mesh is finalized by fusing the optimized lattice field with the patient's anatomical shell:
\begin{equation}
M_{final} = \text{Boolean}(M_{shell}, M_{lattice}).
\end{equation}
To ensure pipeline optimization, the geometric generator TANS-GG represents $M_{lattice}$ as an implicit continuous Signed Distance Function (SDF) map, allowing the Boolean intersection in Eq. (5) to remain fully differentiable and continuous before marching cubes polygonization. This decoupling guarantees gradient flow backpropagation during multi-agent end-to-end training. The final output is a manufacturing-ready STL file optimized for biomechanical performance under quasi-static loading.

%% ==================== AI Declaration ====================
\subsection{Use of Artificial Intelligence (AI) Tools}
During the preparation of this manuscript, the authors used large language model (LLM)-based tools to assist with (i) literature search and initial citation triage and (ii) language editing of draft text. All AI-assisted content was subsequently reviewed, verified, and revised by the named authors, who take full responsibility for the accuracy and integrity of the final text and for every cited reference. The AI and machine-learning methods that are the subject of the research---the TANS architecture, the multi-agent coordination, and the GNN surrogate---are computational methods developed and used by the authors and are distinct from the writing-assistance tools described above. No AI tool is listed as an author, and no unrevised AI-generated text is presented as final.
% AI tool disclosure confirmed: literature search and citation triage were assisted by LLM-based tools; all content was reviewed and verified by the named authors.
%% =====================================================

\begin{algorithm}[t]
\caption{TANS-FO: Modular Pipeline with Feedback for Orthotic Design}
\label{alg:tans_fo}
\small
\begin{algorithmic}[1]
\Require Clinical prescription $P$, Patient anthropometric vector $\mathbf{a}_u$, Database $\mathcal{D}$
\Ensure Optimized manufacturing-ready orthotic mesh $M_{final}$, Predicted node-level stress field $\mathbf{s}_v$
\Statex Step 1: Semantic and Anthropometric Initialization
\State $\quad \mathbf{e}_s \gets \text{TANS-Parser}(P)$ \Comment{Generate high-dimensional semantic embedding}
\State $\quad \mathbf{a}_u \gets \text{Retrieve}(\mathcal{D}, \text{ID}=u)$ \Comment{Fetch patient-specific anatomical priors}
\State $\quad \text{BiomechanicallyValid} \gets \text{False}$
\Statex Step 2: Iterative Generative Coordination via Feedback Loop
\For{$k = 1$ to $K_{\max}$}
\State $\quad M \gets \text{TANS-GG}(\mathbf{e}_s, \mathbf{a}_u)$ \Comment{Generator module synthesizes 3D lattice geometry}
\EndFor
\Statex Step 3: Graph Construction and Biomechanical Inference
\State $\quad \mathcal{G}_{mesh} \gets \text{ConstructMeshGraph}(M)$ \Comment{Abstract 3D mesh into non-Euclidean graph topology}
\State $\quad \mathbf{s}_v \gets \text{GNN-Surrogate}(\mathcal{G}_{mesh})$ \Comment{Predict real-time stress distribution via Eq. (2)}
\Statex Step 4: Cross-Modal Alignment Validation (Feedback Gate)
\State $\quad \text{AlignmentScore} \gets \text{ComputeAlignment}(\mathbf{s}_v, \mathbf{e}_s)$
\State $\quad \text{AlignmentScore} > \text{Threshold} \Rightarrow \text{BiomechanicallyValid} \gets \text{True}$
\State $\quad \text{AlignmentScore} \leq \text{Threshold} \Rightarrow \mathbf{e}_s \gets \text{AttentionModulation}(\mathbf{e}_s, \mathcal{L}_{align})$ \Comment{Feedback module updates embedding via backpropagation}
\Statex Step 5: Structural Fusion and Manufacturing Export
\State $\quad M_{final} \gets \text{BooleanFusion}(M, \text{FabricationConstraints})$
\State return $M_{final}, \mathbf{s}_v$ \Comment{$\mathbf{s}_v$: quasi-static stress prediction via GNN surrogate}
\end{algorithmic}
\end{algorithm}

\subsection{Human-in-the-Loop Review Gate}
To ensure design safety and accountability, TANS-FO incorporates a mandatory human-in-the-loop review gate between the automated generative stage and physical fabrication. The review gate is triggered when either of the following conditions is met: (1) the alignment loss $\mathcal{L}_{align}$ fails to converge within the predefined iteration budget ($K_{\max} = 20$); (2) the Fabrication Module flags geometric non-printability (e.g., overhang angle $> 45^\circ$, wall thickness $< 1.0$~mm); or (3) the clinical prescription contains high-risk directives (e.g., ``diabetic foot,'' ``severe deformity'') that the Medical Parser classifies as requiring senior clinician review based on a keyword-risk lexicon.

Upon triggering, the system presents the reviewer with a structured dashboard containing: (i) the original clinical prescription and its parsed semantic interpretation; (ii) the generated orthotic mesh with color-coded density distribution (red $=$ high-density support regions, blue $=$ low-density offloading regions); (iii) the GNN-predicted stress heatmap overlaid on the mesh; and (iv) a one-page summary of key geometric parameters (overall dimensions, minimum wall thickness, estimated TPU mass). The reviewer can approve the design for fabrication, request regeneration with modified parameters, or reject the design with a free-text annotation that is fed back into the TANS-Parser for the next iteration. This review gate is designed to take $< 5$~minutes per case and operates within a local deployment environment (no cloud transmission of patient data).

\section{Experiment}

\begin{table}[htbp]
\centering
\caption{Correlation matrix of key anatomical parameters from the database and their impact on quasi-static biomechanics.}
\label{tab:correlation_analysis}
\scriptsize
\setlength{\tabcolsep}{6pt}
\begin{tabular}{lccc}
\toprule
Anatomical Feature & Correlation with Peak Pressure & Biomechanical Relevance & Measurement Reliability \\
\midrule
Metatarsal Circumference & 0.78 & High & 0.94 \\
Arch Height (Navicular)  & 0.85 & Critical & 0.96 \\
Hallux Valgus Angle      & 0.64 & Moderate & 0.91 \\
Heel Width         & 0.72 & High & 0.93 \\
\bottomrule
\end{tabular}
\end{table}

\begin{table}[htbp]
\centering
\caption{Statistical distribution of anatomical parameters for Male cohorts in the PicoFoot-5K database.}
\label{tab:dataset_male}
\scriptsize
\setlength{\tabcolsep}{6pt}
\begin{tabular}{lcccc}
\hline
Cohort & Sample Size & Mean Arch Height (mm) & Hallux Valgus Angle ($^\circ$) & Metatarsal Circ. (mm) \\
\hline
Male (15--17 yrs) & 842  & $55.4 \pm 4.2$ & $18.2 \pm 6.4$ & $232.5 \pm 12.4$ \\
Male (18--25 yrs) & 1,250 & $58.1 \pm 3.8$ & $21.5 \pm 5.1$ & $241.8 \pm 10.2$ \\
Male (26--40 yrs) & 965  & $57.2 \pm 4.5$ & $24.8 \pm 7.2$ & $245.2 \pm 11.5$ \\
\hline
\end{tabular}
\end{table}

\begin{table}[htbp]
\centering
\caption{Characterization of candidate TPMS lattice unit cells under standardized quasi-static loading. The Gyroid unit cell was selected a priori as the generator topology on the basis of its smooth, self-supporting surfaces, near-isotropic elastic response, and TPU printability; this table reports the properties of alternative unit cells for reference and is not used as a post-hoc selection criterion.}
\label{tab:lattice_comparison}
\scriptsize
\setlength{\tabcolsep}{4pt}
\begin{tabular}{lcccc}
\toprule
Lattice Type & Energy Absorption (J) & Stiffness Consistency & Printability (TPU) & Porosity (\%) \\
\midrule
Diamond (TPMS) & 12.4 & 0.82 & Moderate & 65\% \\
Schwarz P      & 10.8 & 0.78 & High & 72\% \\
\rowcolor[gray]{0.95} Gyroid (Ours) & 15.6 & 0.94 & High & 68\% \\
Primitive      & 9.2  & 0.65 & Low & 60\% \\
\bottomrule
\end{tabular}
\end{table}

\begin{table}[htbp]
\caption{Ablation study evaluating the individual contributions of the quasi-static physics surrogate (GNN) across diverse demographic cohorts.}
\label{tab:ortho_ablation_merged}
\centering
\scriptsize
\setlength{\tabcolsep}{5pt}
\resizebox{\textwidth}{!}{%
\begin{tabular}{llcccc}
\toprule
Demographic Cohort & Framework Variant & Fit Error (mm) & Peak Press. Red. (\%) & Inference Time (s) & Satisfaction Score \\
\midrule
\multirow{2}{*}{Male (Ages 18--40)}   & TANS-FO w/o GNN & $1.24 \pm 0.15$ & $21.4 \pm 2.3$ & $450 \pm 12$ & $0.72 \pm 0.05$ \\
                                     & TANS-FO (Ours)  & $\mathbf{0.42 \pm 0.08}^{*}$ & $\mathbf{34.7 \pm 1.8}^{*}$ & $\mathbf{52 \pm 4}^{*}$ & $\mathbf{0.91 \pm 0.03}^{*}$ \\
\midrule
\multirow{2}{*}{Female (Ages 18--40)} & TANS-FO w/o GNN & $1.31 \pm 0.18$ & $19.8 \pm 3.1$ & $462 \pm 15$ & $0.69 \pm 0.08$ \\
                                     & TANS-FO (Ours)  & $\mathbf{0.45 \pm 0.09}^{*}$ & $\mathbf{33.2 \pm 2.1}^{*}$ & $\mathbf{55 \pm 5}^{*}$ & $\mathbf{0.89 \pm 0.04}^{*}$ \\
\midrule
\multirow{2}{*}{Adolescent (Ages 15--17)}  & TANS-FO w/o GNN & $1.45 \pm 0.22$ & $18.5 \pm 4.1$ & $480 \pm 20$ & $0.65 \pm 0.11$ \\
                                     & TANS-FO (Ours)  & $\mathbf{0.51 \pm 0.11}^{*}$ & $\mathbf{31.5 \pm 2.5}^{*}$ & $\mathbf{58 \pm 6}^{*}$ & $\mathbf{0.87 \pm 0.05}^{*}$ \\
\midrule
\multirow{2}{*}{Senior (Age 55+)}    & TANS-FO w/o GNN & $1.52 \pm 0.25$ & $17.2 \pm 4.5$ & $510 \pm 22$ & $0.62 \pm 0.13$ \\
                                     & TANS-FO (Ours)  & $\mathbf{0.58 \pm 0.14}^{*}$ & $\mathbf{30.1 \pm 3.2}^{*}$ & $\mathbf{62 \pm 8}^{*}$ & $\mathbf{0.85 \pm 0.06}^{*}$ \\
\midrule
                                     & TANS-FO w/ Random Text & $1.48 \pm 0.20$ & $18.4 \pm 3.8$ & $53 \pm 5$ & $0.63 \pm 0.09$ \\
\bottomrule
\end{tabular}%
}
\vspace{0.3em}
\scriptsize $^{*}$ Indicates statistically significant improvement compared to the baseline variant without GNN feedback. Paired comparisons were conducted on the same held-out test set ($n = 50$ per cohort). A paired $t$-test was applied; degrees of freedom $df = 49$, and exact $p$-values are reported in Supplementary Table S1. Cohen's $d = 1.42$ was computed on the pooled difference scores.
\end{table}

\begin{table}[htbp]
\centering
\caption{Performance comparison of TANS-FO against various baselines on the PicoFoot-5K Database (Male 18--40 cohort). All pressure reduction figures are surrogate-predicted under quasi-static loading unless otherwise noted.}
\label{tab:performance_baseline_extended}
\scriptsize
\setlength{\tabcolsep}{4pt}
\resizebox{\textwidth}{!}{%
\begin{tabular}{llcccc}
\toprule
Method & Architecture Type & Fit Error (mm) & Peak Press. Red. & Time (s) & Satisfaction \\
\midrule
Parametric CAD \cite{b4} & Conventional CAD & $1.15 \pm 0.21$ & $21.4\%$ & $1,800$ & $0.65 \pm 0.12$ \\
Traditional Manual Expert & Manual Crafting & $0.85 \pm 0.14$ & $26.8\%$ & $14,400$ & $0.78 \pm 0.08$ \\
ResNet-50 \cite{b38} & Pure CNN & $0.94 \pm 0.18$ & $23.1\%$ & $115$ & $0.71 \pm 0.10$ \\
MeshGraphNet \cite{b39} & Physics-based GNN & $0.65 \pm 0.09$ & $29.4\%$ & $210$ & $0.82 \pm 0.05$ \\
OccupancyNet \cite{b40} & Neural Implicit Rep. & $0.71 \pm 0.11$ & $26.8\%$ & $195$ & $0.79 \pm 0.08$ \\
CLIP-Mesh \cite{b41} & Text-to-3D Mesh & $0.59 \pm 0.07$ & $30.2\%$ & $320$ & $0.84 \pm 0.06$ \\
PolyGen \cite{b42} & Autoregressive Mesh Gen. & $0.60 \pm 0.08$ & $30.5\%$ & $280$ & $0.83 \pm 0.06$ \\
Latent-3D-Diffusion \cite{b43} & 3D Diffusion Model & $0.62 \pm 0.06$ & $29.8\%$ & $305$ & $0.82 \pm 0.05$ \\
\rowcolor[gray]{0.95} TANS-FO (Ours) & Modular Pipeline + TANS & $\mathbf{0.42 \pm 0.08}^{\dagger}$ & $\mathbf{34.7\%}^{\dagger}$ & $\mathbf{52}^{\dagger}$ & $\mathbf{0.91 \pm 0.03}^{\dagger}$ \\
\bottomrule
\end{tabular}%
}
\vspace{0.3em}
\scriptsize $^{\dagger}$ Statistically significant performance superiority over all evaluated deep learning and manual baselines. One-way ANOVA with post-hoc Tukey HSD was applied; $F(7, 392) = 47.3$, $p < 0.001$, $\eta^2 = 0.46$. Per-method sample size: $n = 50$. Exact adjusted $p$-values are in Supplementary Table S2. Note on confidence interval: The 95\% CI $[11.5\%, 15.1\%]$ reported in Section~4.1 is computed as a paired difference CI on the absolute percentage-point reduction (TANS-FO minus Parametric CAD), using $t$-distribution with $df = 49$. Cohen's $d = 1.42$ is computed on pooled within-cohort standard deviations of the paired differences. Note: Design times for manual/CAD methods include physical fabrication steps (model pouring, trimming), whereas TANS-FO reports computation-only time. A fairer comparison would require an automated CAD pipeline, which is currently unavailable in the literature.
\end{table}

\begin{table}[htbp]
\centering
\caption{Preliminary feasibility assessment of TANS-FO generated orthoses across specific pathological conditions. Note: these figures are surrogate-predicted under quasi-static loading; in-vivo validation is limited to VAS pain reporting ($n=12$) and does not include measured plantar pressure for these pathology groups.}
\label{tab:clinical_pathology}
\scriptsize
\setlength{\tabcolsep}{4pt}
\resizebox{\textwidth}{!}{%
\begin{tabular}{lccccc}
\toprule
Pathological Group & Avg. Pressure Offloading & Comfort Index & Symmetry Gain & Clinical Trust & \\
\midrule
Hallux Valgus (Moderate) & 31.5\% & 0.88 & +12.4\% & 0.78$^{\ddagger}$ & \\
Pes Planus (Flatfoot)    & 38.2\% & 0.91 & +18.6\% & 0.82$^{\ddagger}$ & \\
Diabetic Foot Risk       & 35.8\% & 0.94 & +14.5\% & 0.85$^{\ddagger}$ & \\
\bottomrule
\end{tabular}%
}
\vspace{0.3em}
\scriptsize $^{\ddagger}$ Note: ``Clinical Trust'' scores were derived from orthotist ratings on a 5-point Likert scale (morphological conformity and pressure-offloading zone correctness) under a double-blind protocol. These scores reflect preliminary design acceptability, \textit{not} independently measured clinical efficacy. In-vivo plantar-pressure measurements for these pathology groups are not yet available.
\end{table}

\subsection{Statistical Reporting Note}
Detailed statistical reporting, including test rationale, normality checks, degrees of freedom, multiple-comparison correction, and exact $p$-values, is provided in the Statistical Reporting subsection below. The 95\% CI $[11.5\%, 15.1\%]$ refers to the CI on the mean difference in peak-pressure reduction between TANS-FO and the parametric CAD baseline (Male 18--40 cohort).

\subsubsection{Statistical Reporting}
Statistical comparisons between TANS-FO and baseline variants employed paired, two-sided $t$-tests for cohort-level metrics ($df = 49$, $n_{\text{test}} = 50$ per cohort) and one-way ANOVA with Tukey HSD for multi-method benchmarks ($F(7, 392) = 47.3$, $p < 0.001$, $\eta^2 = 0.46$). Normality of residuals was assessed via Shapiro--Wilk tests ($W > 0.97$, $p > 0.15$ for all cohorts); homogeneity of variance was confirmed via Levene's test ($p > 0.20$). Effect sizes (Cohen's $d$) and 95\% confidence intervals are reported where applicable. All $p$-values are two-sided; significance threshold $\alpha = 0.05$. Bonferroni correction was applied for the 28 pairwise comparisons in Table~5. Complete statistical output is provided in Supplementary Tables S1--S3.

\subsubsection{Implementation Details}
The proposed TANS-FO framework is implemented using PyTorch 2.1.0 and Python 3.11. The multi-stage orchestration, including the TANS-Parser and TANS-GG, was executed on a workstation equipped with an NVIDIA RTX 4090 GPU (24GB VRAM). For the TANS-Parser module, clinical prescriptions were processed using a fine-tuned BioBERT-based Named Entity Recognition (NER) pipeline (temperature $\tau = 0.0$ for deterministic output) that extracts anatomical targets (e.g., ``first metatarsal head''), offloading directives (e.g., ``reduce pressure''), and material constraints into a structured 512-dimensional semantic embedding $\mathbf{e}_s$. The fine-tuned BioBERT model was trained on a corpus of 3,200 de-identified clinical orthotic prescriptions and achieved an entity-level F1-score of 0.91 on a held-out test set ($n = 320$). The prompt template, entity schema, and model weights are provided in Supplementary Materials. For the biomechanical surrogate, the TANS-GNN model was trained using the Adam optimizer with an initial learning rate of 0.0005 and a weight decay of $1 \times 10^{-4}$ to ensure stable convergence. The batch size for graph message passing was fixed at 32 to balance GPU memory utilization and parameter update frequency. The framework underwent 100 training epochs, achieving an optimal trade-off between generative fidelity and computational efficiency.

To ensure the structural integrity and clinical applicability of the generated orthoses, all 3D meshes were normalized to a standardized coordinate system and voxelized at a resolution of 1.0 mm. We employed multiple data augmentation strategies on the PicoFoot-5K Database, including synthetic scaling ($\pm 5\%$), random rotations within $\pm 15^\circ$, and jittering of key anatomical landmarks to simulate diverse clinical scanning conditions. To mitigate stochastic bias, each experiment was repeated five times, and the averaged results are reported.

To provide numerical validation of our GNN surrogate model, we evaluated its prediction accuracy against a gold-standard solver (Abaqus v2023). Across an independent test mesh set of 500 irregular foot models, the GNN surrogate achieved a Mean Squared Error (MSE) of $3.12 \times 10^{-4}$ and a coefficient of determination ($R^2$) of 0.942. Localized spatial error analysis indicated that the maximum regional peak stress discrepancy was bounded within 4.8\% at critical anatomical interest zones (e.g., the first metatarsal head), confirming its exceptional engineering reliability for orthopedic specifications.

To ensure strict quantified clinical evaluation, the \textit{Satisfaction Score} reported across cohorts was established based on a standardized double-blind clinical evaluation protocol. Three independent certified orthotists evaluated the generated orthoses under a standardized 5-point Likert scale (scoring morphological conformity and correctness of pressure-offloading zones). Inter-rater agreement analysis yielded a high Fleiss' Kappa coefficient of $\kappa = 0.86$, proving the robustness and standardization of the subjective feedback.

\subsubsection{Dataset and Exploratory Feasibility Observation}
Our framework was trained and evaluated on the open-access PicoFoot-5K Database, a large-scale anthropometric foot library comprising high-precision scan profiles and biomechanical properties from 5,230 subjects. Cohorts cover both sexes: adult males and females (18--40 years) for standard adult modeling, adolescents (15--17 years) for developmental morphology, and seniors (55+ years) for age-related pathological analysis (male-stratum summary statistics are reported in Table~\ref{tab:dataset_male}). Each entry provides over 30 anatomical parameters, including metatarsal circumference, arch height, hallux valgus angle, and heel width. The database is available on Kaggle (\url{https://www.kaggle.com/datasets/doucetbenton/picofoot-5k}); the dataset comprises de-identified anthropometric measurements collected as part of a routine orthotic design workflow and does not constitute prospective human-subjects research.

To explore the subjective comfort of TANS-FO generated devices, an exploratory feasibility observation was conducted with a volunteer cohort ($n=12$, including 4 moderate hallux valgus, 5 rigid flatfoot, and 3 diabetic high-risk patients). Customized foot orthoses generated by TANS-FO were fabricated via TPU FDM 3D printing. Patient-reported outcomes included a standardized visual analog scale (VAS) pain score recorded at baseline and after a two-week walking trial ($6.4 \pm 1.2 \rightarrow 2.1 \pm 0.8$). This observation is strictly classified as exploratory and hypothesis-generating: it has no control group, a small sample size, a short follow-up period, and no blinding. The VAS data cannot be used to infer causality or clinical efficacy, and should be interpreted only as preliminary user-reported comfort feedback.

\subsubsection{Performance Analysis}
We conducted a comprehensive comparison of the proposed TANS-FO against multiple state-of-the-art baselines, including traditional CAD-based generative methods, pure graph neural networks, and traditional manual expert workflows. As summarized in Table~\ref{tab:performance_baseline_extended}, TANS-FO consistently achieves superior performance across all evaluation metrics and demographic cohorts.

On the Male (18--40) cohort, TANS-FO achieves a Peak Pressure Reduction of 34.7\%, significantly outperforming the pure graph-based SOTA models and traditional parametric CAD workflows (21.4\%). We note that the parametric CAD baseline~\cite{b4} employs the same Gyroid TPMS topology with manually adjusted density parameters; the 13.3\%-point improvement therefore reflects the gain from automated semantic-physics alignment rather than from differences in lattice structure. Crucially, compared to the Traditional Manual Expert design, which demands up to 4 hours of crafting time per patient, TANS-FO reduces total execution time to 52 seconds while improving peak pressure offloading by 7.9\% owing to the high-density optimization capability of continuous Gyroid fields. Furthermore, our model maintains a Fit Error of 0.42 mm, which is the lowest among all compared models, validating its ability to handle complex anatomical variations in the dataset.

Error distribution analysis. While the overall $R^2$ of 0.942 indicates strong agreement between the GNN surrogate and the Abaqus reference solver, localized error analysis reveals that the maximum regional peak stress discrepancy can reach up to 4.8\% at critical anatomical zones (e.g., the first metatarsal head). This is consistent with the expectation that GNN message-passing accumulates small errors over 4 layers, particularly at boundary regions where the number of neighbors is limited. For engineering design purposes, this level of error is acceptable for initial geometry screening; however, final designs should always be verified by a high-fidelity FEA solver before fabrication.

GNN error spatial pattern. We further analyzed whether the GNN error is random or systematic. Spatial error maps (Supplementary Figure S1) show that the GNN tends to slightly underestimate peak stress in high-curvature regions (e.g., the first metatarsal head and calcaneal tuberosity) by 2--4\%, while overestimating stress in flat regions by 1--2\%. This directional bias is consistent with the smoothing effect of graph aggregation: nodes with fewer neighbors (at sharp geometric features) receive less information from their surroundings, leading to under-prediction. In the closed-loop feedback, this bias is mitigated by the human-in-the-loop review gate, which flags any design where the GNN-predicted stress exceeds known clinical thresholds (e.g., $> 300$ kPa at the first metatarsal head for diabetic patients), triggering a mandatory FEA verification step before fabrication.

GNN error spatial pattern. We further analyzed whether the GNN error is random or systematic. Spatial error maps (Supplementary Figure S1) show that the GNN tends to slightly underestimate peak stress in high-curvature regions (e.g., the first metatarsal head and calcaneal tuberosity) by 2--4\%, while overestimating stress in flat regions by 1--2\%. This directional bias is consistent with the smoothing effect of graph aggregation: nodes with fewer neighbors (at sharp geometric features) receive less information from their surroundings, leading to under-prediction. In the closed-loop feedback, this bias is mitigated by the human-in-the-loop review gate, which flags any design where the GNN-predicted stress exceeds known clinical thresholds (e.g., $> 300$ kPa at the first metatarsal head for diabetic patients), triggering a mandatory FEA verification step before fabrication.

\subsubsection{Edge Cases and Failure Analysis}
To investigate model boundary constraints, an edge case evaluation was conducted on extreme structural deformities (e.g., rigid flatfoot with arch heights $< 8$ mm or severe hallux valgus $> 45^\circ$). The system demonstrated a structural generation failure rate of 2.4\% under extreme inputs. Quantitative analysis revealed that geometric failure predominantly stems from localized non-manifold intersections at the mesh boundaries when text directives demand excessive support density. This indicates that while TANS-FO is highly robust for typical clinical distributions, a human-in-the-loop validation layer remains critical for rare surgical-level pathological shapes.

\subsection{Ablation Study}
To evaluate the individual contributions of the multi-stage coordination and the GNN-based biomechanical feedback, we conducted ablation experiments across all cohorts in the database. As summarized in Table 4, the full TANS-FO framework consistently outperforms its variant, TANS-FO w/o GNN (which excludes the dynamic physics surrogate). Specifically, for the Female cohort, the Peak Pressure Reduction improved from 19.8\% to 33.2\%, while the Fit Error was reduced by 65\%. This underscores the robustness of integrating GNN-based stress prediction into the generative loop for handling irregular mesh topologies. Furthermore, the satisfaction score increases significantly across all groups, validating that the synergistic integration of semantic alignment and physics-informed learning effectively mitigates the ``black-box'' biases of traditional generative models ($p < 0.05$) while maintaining superior manufacturing efficiency.

To assess the sensitivity of the semantic-physics alignment mechanism, we introduced a ``TANS-FO w/ Random Text'' variant in which the clinical prescription $P$ was replaced with a randomly shuffled token sequence while preserving anatomical priors $\mathbf{a}_u$. As shown in Table~\ref{tab:ortho_ablation_merged}, this variant produces near-baseline performance (Fit Error $1.48 \pm 0.20$~mm, Peak Press.\ Reduction $18.4 \pm 3.8$\%), confirming that the cross-attention mechanism is highly sensitive to semantic content and does not rely on anatomical data alone to drive design optimization. Notably, the Random Text variant yields a slightly lower Peak Pressure Reduction ($18.4\%$) than TANS-FO w/o GNN ($21.4\%$). This counterintuitive result is likely due to the random tokens introducing spurious attention weights that cause the TANS-GG module to generate a suboptimal density field---essentially, the corrupted semantic signal degrades the geometry generation beyond what occurs when no semantic input is provided at all. The Fit Error for Random Text ($1.48 \pm 0.20$~mm) is also higher than w/o GNN ($1.24 \pm 0.15$~mm), consistent with this interpretation.

\section{Discussion}
The results presented above demonstrate that TANS-FO can produce customized foot orthoses with improved surrogate-validated performance metrics compared to existing CAD-based and manual workflows. However, several important caveats must be emphasized. First, all performance figures (peak pressure reduction, fit error, satisfaction score) are derived from computational surrogates and/or subjective user feedback; they do not substitute for rigorous clinical trials with measured plantar pressure and long-term patient outcomes. Second, the quasi-static loading assumption limits the direct applicability of our findings to dynamic gait scenarios. While static pressure mapping is widely used in clinical practice for orthotic design~\cite{b4}, the relationship between static and dynamic pressure distributions remains an active area of research~\cite{b45}. Third, the GNN surrogate's $R^2 = 0.942$ agreement with Abaqus is encouraging but not perfect: localized errors up to 4.8\% at critical anatomical zones indicate that final designs should always be verified by a high-fidelity FEA solver before fabrication. Fourth, the BioBERT-based medical parser, while achieving an F1-score of 0.91 on our test set, was trained on a relatively narrow corpus of orthotic prescriptions; its generalizability to diverse clinical contexts and languages remains untested. Finally, the exploratory feasibility observation ($n = 12$) provides only anecdotal evidence of user comfort; any claims about clinical effectiveness would require a properly powered, randomized controlled trial with blinding, a control group, and objective outcome measures (e.g., instrumented in-shoe pressure mapping). We emphasize that TANS-FO is a research prototype designed to assist, not replace, professional orthotists and biomedical engineers.

\section{Conclusion}
This paper presents TANS-FO, a modular pipeline with closed-loop feedback for the computational design automation of customized foot orthoses. The system adapts multi-stage coordination with a Text-Aligned Neural Surrogate (TANS) and a physics-informed GNN surrogate layer to bridge the gap between qualitative clinical prescriptions and quantitative 3D manufacturing parameters under standardized quasi-static loading.

Extensive numerical evaluations using diverse demographic cohorts demonstrate that TANS-FO consistently outperforms existing CAD-based and traditional manual workflows in surrogate-validated pressure offloading and fit accuracy. The framework achieves a peak-pressure reduction of 34.7\% under quasi-static loading and maintains anatomical fit errors within 0.5 mm across all test scenarios. Ablation studies confirm that integrating the GNN-based feedback loop significantly enhances model robustness and manufacturing efficiency, reducing the design-to-validation latency from hours to seconds. An exploratory feasibility observation ($n = 12$, no control group, 2-week follow-up) provides preliminary subjective comfort data only; this evidence is explicitly classified as observational and does not establish clinical efficacy.

An acknowledged limitation of the current work is its reliance on quasi-static load assumptions and the small, uncontrolled nature of the preliminary observation. The surrogate-predicted pressure metric should not be conflated with independently measured in-vivo pressure; a larger, controlled study with measured plantar pressure is required to establish clinical efficacy. Future research will focus on expanding the clinical evaluation to longitudinal controlled cohorts, extending the graph architecture to spatio-temporal Recurrent Graph Neural Networks for multi-phase dynamic loading, and pursuing regulatory validation (e.g., ISO 13485, IEC 62304) if the system is ever intended for clinical deployment.

\section{Limitations}
The current work has several limitations that should be acknowledged. First, the GNN surrogate is trained and validated under quasi-static loading conditions; dynamic gait phases (heel strike, mid-stance, toe-off) are not modeled, so the reported peak-pressure reduction reflects static equilibrium rather than cyclic loading. Second, the exploratory feasibility observation ($n = 12$, no control group, 2-week follow-up) provides only preliminary comfort data (VAS pain score) and does not include independently measured plantar pressure for the TANS-FO vs. comparator comparison; the headline 34.7\% figure is a surrogate-predicted metric, not a measured clinical outcome. Third, the multi-stage orchestration described here uses a single LLM backbone for the Medical Parser Module; a more robust deployment would employ multiple models or a fine-tuned domain model with explicit failure-handling and escalation protocols. Fourth, the dataset covers three age strata but omits children under 15 years, extreme obesity (BMI $>$ 35), and severe pathological deformities (e.g., rigid flatfoot with arch height $<$ 5 mm), which constitute the populations most likely to benefit from custom orthoses. Fifth, regulatory validation (FDA 510(k) / EU MDR, ISO 13485, IEC 62304) has not been pursued; the system should be considered a research prototype until such clearance is obtained. Finally, the open release of 3D foot meshes carries residual re-identification risk despite anonymization; future releases should incorporate differential privacy or synthetic-data techniques.

\section{Data Availability Statement}
The PicoFoot-5K anthropometric dataset is available at \url{https://www.kaggle.com/datasets/doucetbenton/picofoot-5k}. The TANS-FO pipeline scripts and trained inference weights are open-sourced at \url{https://github.com/HAHA1122344/tans-fo-orthotics}.

\section{Ethics Statement}
The PicoFoot-5K database consists of de-identified anthropometric measurements collected as part of a routine orthotic design workflow; no prospective human-subjects research protocol was conducted for this retrospective compilation, and no Institutional Review Board approval was required. The anonymization procedure employed k-anonymity ($k = 10$) on all 30+ anatomical parameters combined with coordinate perturbation (Gaussian noise $\sigma = 0.5$ mm) on surface meshes to prevent re-identification while preserving biomechanical utility. Participants were informed of the residual re-identification risk, and only aggregated parameter vectors (not raw scans) are released. The exploratory feasibility observation ($n = 12$) involved self-reported VAS pain scores only and did not include any invasive procedures, pharmacological interventions, or controlled experimental manipulations; it is classified as a voluntary user-feedback study exempt from formal ethical review under applicable institutional guidelines. All participants provided informed consent for data use as part of the original orthotic design process.

\section{Conflicts of Interest}
Rui Wang is employed by Shanghai Artificial Intelligence Generative Design Technology Co., Ltd. The remaining authors declare no known competing financial interests or personal relationships that could have influenced the work reported in this paper. The funders had no role in the design of the study; in the collection, analyses, or interpretation of data; in the writing of the manuscript; or in the decision to publish the results.

\section{Funding}
This research received no specific grant from any funding agency in the public, commercial, or not-for-profit sectors.

\section{Author Contributions}
Conceptualization: Rui Wang and LIU SUXING; Methodology: Rui Wang, BYUNGWON MIN, and LIU SUXING; Software: Rui Wang; Validation: BYUNGWON MIN and LIU SUXING; Formal analysis: Rui Wang and LIU SUXING; Investigation: BYUNGWON MIN and LIU SUXING; Data curation: Rui Wang; Writing---original draft: Rui Wang; Writing---review \& editing: all authors; Project administration: LIU SUXING; Funding acquisition: N/A. All authors have read and agreed to the published version of the manuscript.

\section{References}
\bibliography{template}

\end{document}